\documentclass[letterpaper, 10 pt, conference]{ieeeconf}
\IEEEoverridecommandlockouts
\usepackage{bobbystyle_IEEEjournal}

\usepackage{amsmath,amsfonts}
\usepackage{algorithmic}
\usepackage{algorithm}
\usepackage{array}
\usepackage[caption=false,font=normalsize,labelfont=sf,textfont=sf]{subfig}
\usepackage{textcomp}
\usepackage{stfloats}
\usepackage{url}
\usepackage{verbatim}
\usepackage{graphicx}
\usepackage{cite}
\usepackage{multirow}
\usepackage{ulem}
\makeatletter
\let\NAT@parse\undefined
\makeatother
\usepackage{hyperref}
\hypersetup{
    colorlinks=true,
    linkcolor=blue,
    filecolor=blue,      
    urlcolor=blue,
    citecolor=blue
}

\usepackage[T1]{fontenc}
\usepackage[utf8]{inputenc}

\newcommand{\dquotes}[1]{``#1''}
\usepackage{chngcntr}
\counterwithout{equation}{section}

\newcolumntype{C}[1]{>{\centering\let\newline\\\arraybackslash\hspace{0pt}}m{#1}}  

\newcolumntype{K}[1]{>{\centering\arraybackslash}p{#1}}

\renewcommand{\theequation}{\arabic{equation}}

  {\list{}{\leftmargin=#1\rightmargin=#1}\item[]}%
  {\endlist}
\usepackage{lipsum} 

\usepackage{titlesec}
\titlespacing{\section}{0pt}{*1.0}{*1.0} 
\titlespacing{\subsection}{0pt}{*0.2}{*0.2}
\title{\LARGE \bf SSIL: Self-Supervised Imitation Learning for End-to-End Driving}

\author{Jin Bok Park, Jinkyu Lee, Muhyun Back, Hyun Min Han, \\ Tianwei Ma, Sang Min Won$^{\dagger}$, Sung Soo Hwang$^{\dagger}$, Il Yong Chun$^{\dagger}$%
\thanks{
$^\dag$Corresponding authors.
The work was supported in part by NRF Grant RS-2023-00213455, RS-2025-24683458, and RS-2026-25480389 funded by MSIT, 
the Digital Therapeutics Development and Demonstration Support Program Grant H0601-24-1023 funded by MSIT and NIPA, 
the BK21 FOUR Project, 
IITP Grant RS-2019-II190421 (AI Graduate School Support Program (Sungkyunkwan University)) funded by MSIT, 
KIAT Grant RS-2024-00418086 (HRD Program for Industrial Innovation) funded by MOTIE, and IBS-R015-D1.
}
\thanks{Jin Bok Park is with the R~\&~D Center, SL Corporation, Anyang, South korea. He was with the Department of Electrical and Computer Engineering, Sungkyunkwan University, Suwon 16419, South Korea (email: bjb663@g.skku.edu).}
\thanks{Jinkyu Lee, Muhyun Back, and Hyun Min Han are with the Department of Information and Communication Engineering, Handong Global University, Pohang, 37554, South Korea (e-mails: jinkyu.lee@stradvision.com; 21931005@handong.edu; manduss1204@gmail.com).}
\thanks{
Tianwei Ma is with the College of Engineering, Texas A~\&~M University-Corpus Christi, Corpus Christi, TX 78412 USA(e-mail: david.ma@tamucc.edu).
}
\thanks{Sang Min Won are with the Department of Electrical and Computer Engineering, Sungkyunkwan University, Suwon, 16419, South Korea (e-mail: sangminwon@skku.edu).}
\thanks{
Sung Soo Hwang is with the School of Computer Science and Electrical Engineering, Handong Global University, Pohang, 37554, South Korea 
(e-mail: sshwang@handong.edu).
}
\thanks{Il Yong Chun is with the Departments of Electrical and Computer Engineering, Artificial Intelligence, Advanced Display Engineering, Display Convergence Engineering, and Semiconductor Convergence Engineering, Sungkyunkwan University, Suwon 16419, South Korea, and also with the Center for Neuroscience Imaging Research, Institute for Basic Science, Suwon 16419, South Korea (email: iychun@skku.edu).}
}
\begin{document}
\maketitle

\begin{abstract}
In autonomous driving, the end-to-end (E2E) driving approach that predicts vehicle control signals directly from sensor data is rapidly gaining attention. 
To learn a safe E2E driving system, one needs an extensive amount of driving data and human intervention.
Vehicle control data is constructed by many hours of human driving, and it is challenging to construct large vehicle control datasets. 
Often, publicly available driving datasets are collected with limited driving scenes, and collecting vehicle control data is only available by vehicle manufacturers.
To address these challenges, this paper proposes the first self-supervised learning framework, \emph{S}elf-\emph{S}upervised \emph{I}mitation \emph{L}earning (\emph{SSIL}), for E2E driving.
The proposed SSIL framework can learn vision-based E2E driving networks without using driving command data or a pre-trained model. 
To construct pseudo steering angle data, proposed SSIL predicts a pseudo target from the vehicle's poses at the current and previous time points that are estimated with light detection and ranging sensors.
In addition, we propose a new cross-attention-based conditioning approach (CACA) for a vision encoder in E2E driving, where a high-level instruction serves as the conditioning signal for visual information.
Our numerical experiments with three different benchmark datasets demonstrate that the proposed SSIL framework achieves \textit{very} comparable E2E driving accuracy with the supervised learning counterpart.
Furthermore, 
the proposed pseudo-label predictor outperformed an existing one using proportional integral derivative controller, and proposed CACA achieved superior performance over existing conditioning approaches.
\end{abstract}

\maketitle
\vspace{1em}
\section{Introduction}
\label{sec:intro}

End-to-end (E2E) driving predicts driving commands (at one end) from sensory input (at the other end), and it is gaining rapid attention for autonomous driving vehicles \cite{Survey_E2E_Driving_Architectures_TrainingMethods}.
The conventional modular approach accomplishes complex tasks of autonomous driving by using interconnected modules such as perception (e.g., object detection), localization, path planning, and control \cite{Survey_E2E_Driving_Architectures_TrainingMethods}.
However, the conventional modular approach needs tremendous efforts to build, maintain, and optimize their modules.

Different from the modular approach, the E2E approach considers the entire processing and prediction pipeline as a single learnable machine-learning task \cite{PilotNet, Transfuser_1, Transfuser_2,E2E_Conditional_IL,E2E_Conditional_IL_2}.
The approach \textit{directly} transforms sensory inputs to driving commands, such as steering angle, acceleration, and braking, \textit{without} additional post-processing, e.g., proportional integral derivative (PID) controller in future trajectory estimation.
Broadly speaking, an E2E driving model can be trained with the two approaches, imitation learning \cite{PilotNet, Transfuser_1, Transfuser_2,E2E_Conditional_IL} and reinforcement learning \cite{Simulation_Based_Reinforcement_Learning, E2E_Reinforce_UrbanDrivingPolicies}.
The former approach learns E2E driving models to mimic human driver behavior.
The latter approach learns E2E driving models to act optimally at each instant.

Both approaches require human driving commands, the final output of E2E driving systems.
Conventionally, imitation learning is supervised learning that requires (labeled) reference data, e.g., steering wheel angle control by human drivers \cite{PilotNet}.
Reinforcement learning requires a significant amount of training data generally with simulators \cite{Simulation_Based_Reinforcement_Learning, E2E_Reinforce_UrbanDrivingPolicies}.
To develop accurate and reliable E2E driving models, 
it is critical to construct a large dataset with many pairs of sensory data and expert driving commands,
particularly collected from diverse driving environments.

Constructing a large driving command dataset is \textit{non}-trivial.
There exist several publicly available datasets for learning autonomous driving systems that provide data collected from sensors mounted on the exterior of vehicles \cite{a2d2dataset,nuScenes,CARLA}.
On the contrary, driving command data is \textit{not} always available \cite{Cityscapes,RobotCar,ApolloScape}.
It is extremely challenging to curate driving command data, because one cannot access internal driving command data such as steering angle without the assistance of vehicle manufacturers.

\begin{figure*}[!tp]
\centering
\includegraphics[width=5in,trim={0.01cm 0.03cm 0.01cm 0.01cm},clip]
{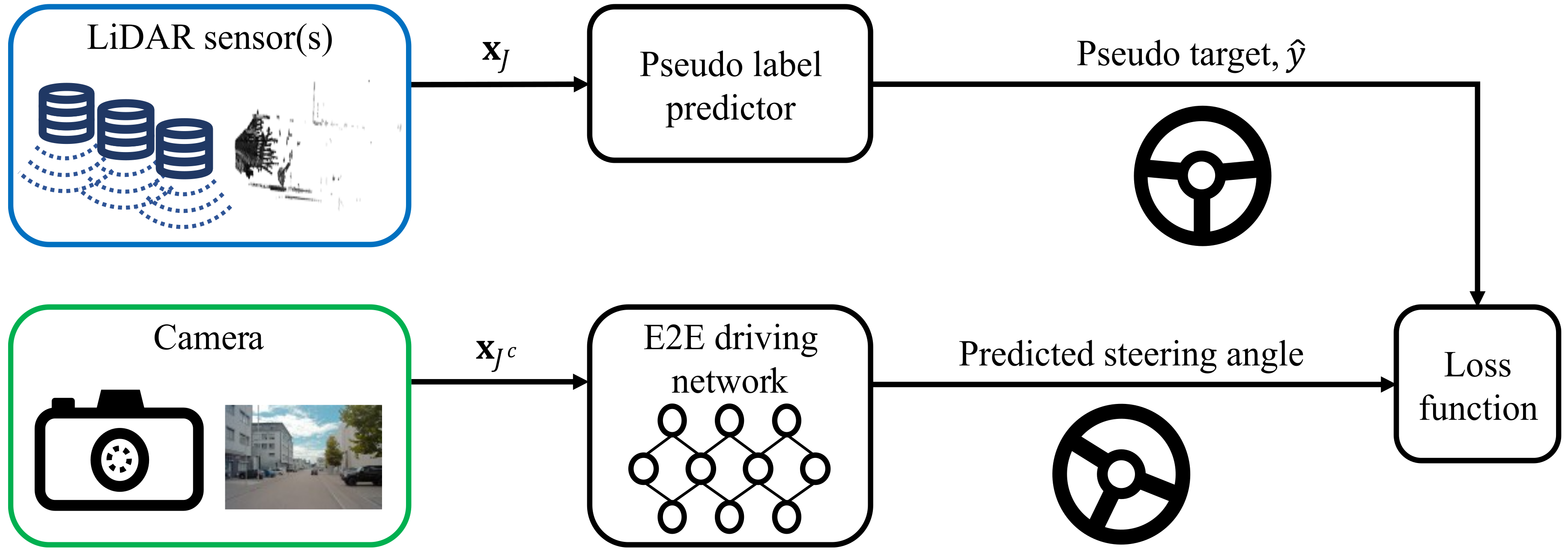}
\vspace{-0.5pc}
\caption{Overview of the proposed SSIL framework.
We modify the general SSRL framework \cite{Chun&etal:22arxiv} by designing a pseudo-label predictor using domain knowledge. 
We designed it as the composition of a vehicle pose estimator using point clouds $\mb{x}_{J}$ generated from LiDAR sensor(s), and a pseudo steering angle predictor.
In comparing SSIL with the ordinary supervised learning method, 
we tested two E2E driving network architectures with three different conditioning approaches using high-level instruction (including proposed CACM).
To train an E2E driving network that takes a camera image $\mb{x}_{J^c}$ as input and predicts a steering angle, a loss function measures the discrepancy between pseudo target $\hat{y}$ and the predicted steering angle.
}
\label{fig:SSIL}
\vspace*{-0.2in}
\end{figure*}

To resolve the aforementioned challenges to develop accurate and reliable E2E driving systems,
one may instead estimate steering angles from sensory data.
This paper proposes the first self-supervised learning framework for vision/camera-based E2E driving, referred to as \textbf{S}elf-\textbf{S}upervised \textbf{I}mitation \textbf{L}earning (\textbf{SSIL}).
We modify the existing self-supervised regression learning (SSRL) approach \cite{Chun&etal:22arxiv}, using domain knowledge in range data captured by light detection and ranging (LiDAR) sensor(s), vehicle geometry, steering geometry \cite{Vehicle_dynamics}, etc.
To construct pseudo steering angle data, we estimate the vehicle pose for each time frame from point clouds obtained by LiDAR sensor(s), and predict a pseudo-label at each time point, using the estimated poses from two consecutive time frames and domain knowledge regarding the vehicle.
Furthermore, we propose a new cross-attention-based conditioning approach (CACA) for a vision encoder in E2E driving, which leverages a high-level instruction to guide vision information in predicting a driving command.
Fig.~\ref{fig:SSIL} overviews the proposed SSIL framework.

Our contributions can be summarized as follows:
\begin{itemize}
\item We propose the \emph{first} self-supervised learning framework for E2E driving, based on the SSRL framework.
The proposed SSIL framework can learn high-quality vision-based E2E driving NNs \textit{without} expert driving commands or a pre-trained model, using domain knowledge in range data captured via LiDAR sensor(s), vehicle geometry, and steering geometry.

\item We propose a new conditioning approach for a vision encoder in E2E driving, CACA, that uses the cross-attention mechanism \cite{ViT} to condition visual information with a high-level instruction.



\item Our numerical experiments with three different benchmarks using different vehicles and LiDAR sensors, consistently show that the proposed SSIL framework achieves \textit{very} comparable E2E driving performances with supervised imitation learning (SIL).
The proposed pseudo-label predictor outperformed the existing PID-based one introduced in \cite{Transfuser_1, Transfuser_2}.
Finally, E2E driving NNs using proposed CACA achieved significantly better driving performances than those using the existing conditioning approaches with module selection \cite{E2E_Conditional_IL, E2E_Conditional_IL_2} and self-attention mechanisms \cite{CIL++}.

\end{itemize}

\section{Related works}
\label{sec:related works}

This section reviews related existing E2E driving methods that learn NNs using expert driving commands.
PilotNet is a vision-based E2E driving method with CNN to predict steering angles \cite{PilotNet}.
Conditional imitation learning (CIL) is a vision-based E2E driving method that uses a vision encoder and a conditioning module with high-level instructions (e.g., \dquotes{turn right}, \dquotes{turn left}, and \dquotes{go straight}) to predict driving commands, 
where high-level instructions are assumed given \cite{E2E_Conditional_IL,E2E_Conditional_IL_2}.
The works \cite{E2E_Conditional_IL,E2E_Conditional_IL_2} 
use a module selection mechanism: 
they select a driving command estimator -- implemented as a multi-layer perceptron (MLP) -- depending on a high-level instruction.
The work \cite{CIL++} uses multi-view cameras inputs and a self-attention-based conditioning approach with high-level instructions \cite{CIL++}.
Multi-modal fusion transformer (TransFuser) is a multi-modal E2E driving method that uses an RGB image from a camera and two-channel BEV from a LiDAR sensor as two inputs \cite{Transfuser_1}.
TransFuser uses the attention mechanism of transformer to capture the global context of driving scenes.
Latent TransFuser is a vision-based TransFuser \cite{Transfuser_1}
that replaces two-channel BEV input from a LiDAR sensor with two-channel positional encoding \cite{Transfuser_2}.

To develop safe and reliable E2E driving systems, the aforementioned existing E2E driving methods require extensive driving datasets including driving commands collected by human drivers or expert systems \cite{PilotNet,E2E_Conditional_IL,E2E_Conditional_IL_2,Transfuser_1,Transfuser_2}.
There exist semi-supervised learning methods that can produce a pseudo future trajectory with a pre-trained NN \cite{ACO, SelfD}.
It is reported even with a pre-trained model that 
their performances in producing pseudo driving commands with post-processing are poor.
Different from the aforementioned methods, our proposed SSIL framework can learn E2E driving NNs \textit{without} expert driving commands or a pre-trained model.

\section{Background: SSRL}
\label{sec:ssrl}

SSRL is the first general fully self-supervised learning framework for regression models that predicts continuous quantity and has been successfully applied to computational imaging \cite{Chun&etal:22arxiv}.\footnote{Similar to the conventional self-supervised feature learning concept, 
SSRL uses a masking scheme that divides an input sample into two parts. See \R{sys:ssrl}.}
To learn a regression network $f$ only from input data $\mb{x} \in \bbR^N$ but without ground-truth target data $\mb{y} \in \bbR^M$,
the framework uses a designable \dquotes{pseudo-predictor} $g$ that encapsulates domain knowledge of a specific application:
it minimizes the mean square error (MSE) between predictions from $f$ and $g$ using only input data.
Specifically, the self-supervised regression loss function is given by \cite{Chun&etal:22arxiv}
\be{
\label{sys:ssrl}
\bbE_{\mb{x}} \| f(\mb{x}_{J^c}) - g(\mb{x}_J) \|_2^2,
}
where regression network $f$ and pseudo-predictor $g$ use complementary information of $\mb{x}$, $\mb{x}_{J^c}$ and $\mb{x}_{J}$, respectively,
of which
 $J \in \cJ$, $\cJ$ is a partition of $\{ 1,\ldots,N \}$, $J^c$ denotes the complement of $J$, and $(\cdot)_J$ denotes a vector restricted to $J$.
Remark that $f$ obtained by minimizing \R{sys:ssrl} cannot merely be $g$.
The following properties underscore the importance of designing \dquotes{good} $g$ in the SSRL loss \R{sys:ssrl}:

\thm{[Expected prediction error \cite{Chun&etal:22arxiv}]
\label{thm:ssrl}
Suppose that $f$ and $g$ are measurable.
Then the expected prediction error of the optimal solution $f^\star$ of SSRL \R{sys:ssrl}, $f^\star (\mb{x}) = \bbE[ g(\mb{x}_J) | \mb{x}_{J^c} ]$ for each $J \in \cJ$, at an unseen input $\mb{x}'$ is given by
\be{
\label{eq:thm:soln:err}
\bbE [ \| f^\star (\mb{x}) - \mb{y} \|_2^2 | \mb{x} \!=\! \mb{x}' ]
= \| f^\star (\mb{x}') - f^\ast (\mb{x}') \|_2^2 + \mathrm{Var}( \mb{y} | \mb{x} \!=\! \mb{x}' ),
}
where $f^\ast$ is the optimal solution of the supervision counterpart, $f^\ast (\mb{x}) = \bbE[\mb{y}|\mb{x}_{J^c}]$.
}

The theorem above indicates that the better $g$, 
i.e., the closer $g(\mb{x}_J)$ is to $\mb{y}$,
the better $f^\star (\mb{x}_{J^c})$,
i.e., the closer $f^\star (\mb{x}_{J^c})$ is to the optimal solution of the supervised counterpart, $\bbE_{\mb{x},\mb{y}} \| f(\mb{x}_{J^c}) - \mb{y} \|_2^2 $.
Consequently, such $g$ can diminish the expected prediction error \R{eq:thm:soln:err}, by reducing the first term in \R{eq:thm:soln:err}. 
(Note that the second term indicates an irreducible error.)

In computational imaging applications, the SSRL framework \cite{Chun&etal:22arxiv} uses noise properties in $\mb{x}$ as domain knowledge to design a good pseudo-predictor $g$.
In applying SSIL to E2E driving, 
we use an image collected from a camera as input to an E2E driving NN $f$ conditioned on a high-level instruction. 
We use point clouds collected from LiDAR sensor(s) as input to a pseudo-predictor $g$.
Our aim is to design a good $g$ by using some domain knowledge about LiDAR sensors and vehicle/steering geometry \cite{Vehicle_dynamics}.

\section{Methods: SSIL using camera and LiDAR data}
\label{sec:ssil}

In E2E driving, SIL trains a regression neural network by using pairs of visual input(s) and expert driving command(s) \cite{PilotNet}.
Different from SIL, the proposed SSIL framework aims to train an E2E driving NN \textit{without} using vehicle control data or a pre-trained model.
This section sophisticatedly customizes the the self-supervised learning framework framework in Section~\ref{sec:ssrl} for vision-based E2E driving. 
We will use well-known sensors in autonomous driving, specifically, a camera and LiDAR sensor(s).

In the SSRL loss \R{sys:ssrl}, we first construct an input $\mb{x}$ using
an image collected from a camera, $\mb{x}_{J^c}$,
and point clouds collected from LiDAR sensor(s), $\mb{x}_{J}$:\footnote{
In \R{sys:ssrl}, if two additional assumptions are satisfied,
one can obtain the optimal solution in Theorem~\ref{thm:ssrl} that vanishes the first term in \R{eq:thm:soln:err} \cite{Chun&etal:22arxiv}. 
The constructed setup in \R{eq:3} naturally satisfies the assumption, $p(\mb{x} | \mb{y}) = p(\mb{x}_{J^c} | \mb{y}) \cdot p(\mb{x}_{J} | \mb{y})$, 
because noises in different sensors are statistically independent.
}
\be{
\label{eq:3}
\mb{x} = 
\begin{bmatrix}
 \mb{x}_{J^{c}}\\\mb{x}_{J}
\end{bmatrix},
}
where $\mb{x}$ is collected at the $i$th time point.
In using \R{eq:3},
it is most essential to design a pseudo target predictor $g$ based on application specific knowledge.
We design a pseudo target predictor $g$ as the composition of two operators:
\be{
\label{sys:g_function}
g = {h}_{2}\cdot{h}_{1},
}
where ${h}_{1}$ denotes an operator that estimates two vehicle poses from adjacent time points with $\mb{x}_{J}$, 
and ${h}_{2}$ denotes an operator that predicts a pseudo target steering angle using two adjacent vehicle poses estimated from ${h}_{1}$.

The next three subsections describe the details of $h_1$, $h_2$, and two E2E driving networks $f$.

\subsection{${h}_{1}$: Function generating vehicle poses from LiDAR odometry and mapping}
\label{sec:h1}

In general, 
localizing an ego vehicle with IMU(s) or global positioning system (GPS) is \textit{not} sufficiently accurate.
In many autonomous driving and advanced driver assistance systems,
it is recommended to use some simultaneous localization and mapping methods for vehicle localization.
Using the LiDAR odometry and mapping (LOAM) method \cite{LOAM},
we design the function ${h}_{1}$ to estimate the vehicle poses at the current and previous time points.
LOAM is a real-time algorithm that calculates odometry and mapping using data acquired from LiDAR; A-LOAM is an advanced implementation of LOAM that simplifies the code structure of LOAM.
A-LOAM can perform robust vehicle pose estimation using scan matching and feature extraction, even in
urban driving environments (where tall buildings obstruct GPS signals) and long-term driving scenarios \cite{LOAM}.

\begin{figure}[t!]
\centerline{\hbox{
\includegraphics[width=2.8in,trim={0.01cm 0.2cm 0.01cm 0.1cm},clip]{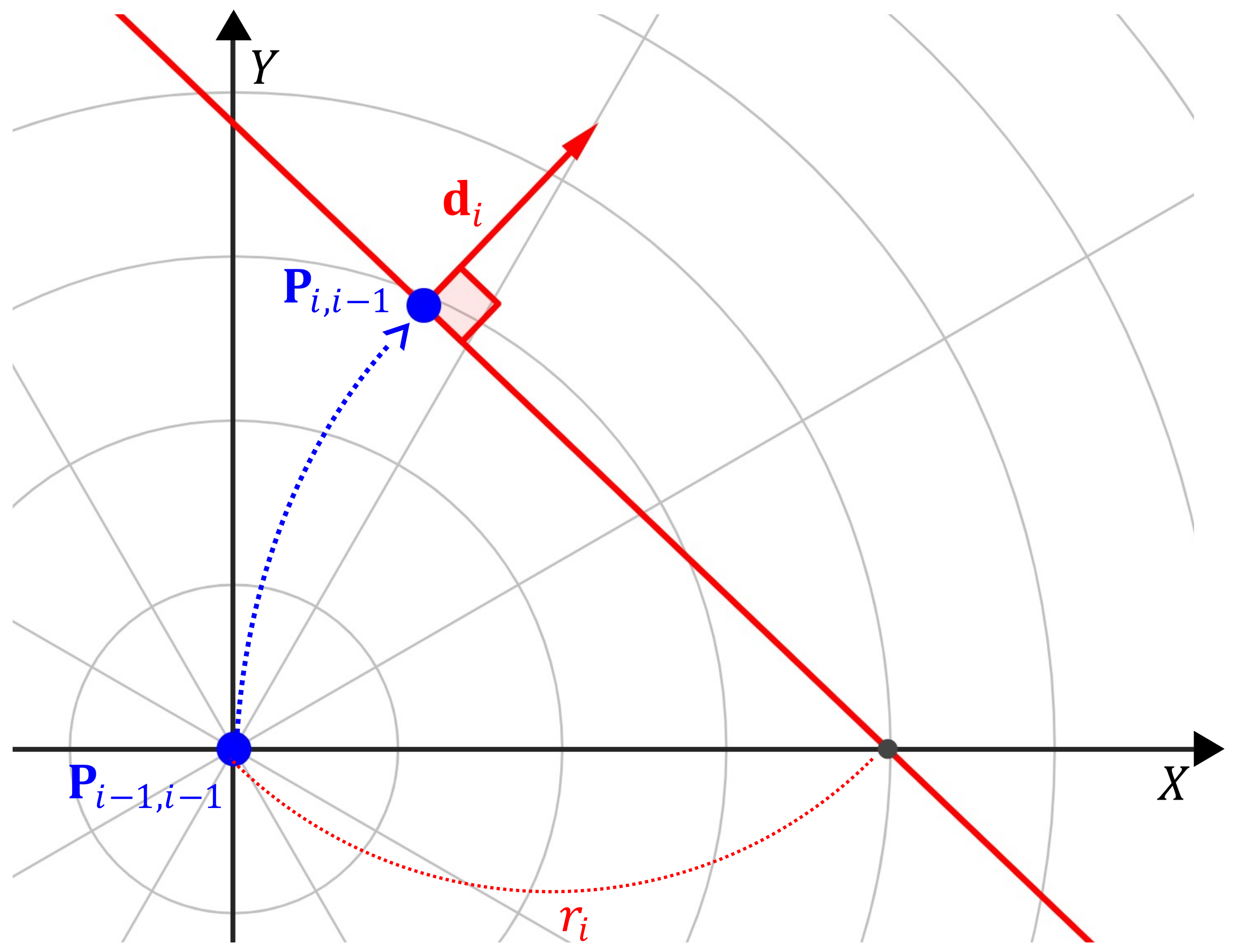}
}}
\vspace{-0.5pc}
\caption{Geometric illustration of calculating the turning radius in the $X$-$Y$ vehicle coordinate system where $\mb{P}_{i-1,i-1}$$ = \mb{I}$.
The matrices $\mb{P}_{i,i-1}$ and $\mb{P}_{i-1,i-1}$ indicate the vehicle poses at the current and previous time points, respectively. 
The vector $\mb{d}_i$ indicates the forward direction vector of the vehicle at the current time point.
The distance $r_i$ is a turning radius of the vehicle at the $i$th time point. 
It is given by the $X$-intercept of the line that is perpendicular to $\mb{d}_i$ and crosses $\mb{t}_{i,i-1}$ (the translation vector of $\mb{P}_{i,i-1}$) in the $X$-$Y$ vehicle coordinate system.
The blue dotted arrow indicates a driving trajectory from the $(i-1)$th to $i$th time point.
}
\label{fig:turning radius}
\vspace*{-0.2in}
\end{figure}

The vehicle pose at the $i$th time point relative to the initial (i.e., $0$th) vehicle pose can be written as follows: 
\be{
\label{sys:vehicle's position}
\mb{P}_{{i},{0}} 
= 
\begin{bmatrix}
\mb{R}_{i,0} & \mb{t}_{i,0}
\\ 
\mb{0}^{\top} & 1
\end{bmatrix},
}
assuming that $\mb{P}_0$ is the initial vehicle pose in the world coordinate system.
(If the initial vehicle position is at the origin of the world coordinate system with no rotation, $\mb{P}_0 = \mb{I}$.)
Here, we estimate ${\mb{P}}_{{i},{0}}$ via A-LOAM, 
and ${{\mb{R}}_{{i},{0}}} \in \bbR^{3 \times 3}$ and ${{\mb{t}}_{{i},{0}}} \in \bbR^{3}$ denote its corresponding rotation matrix and translation vector, respectively.

The next section estimates a steering angle for each time point using \R{sys:vehicle's position} and some additional domain knowledge.

\subsection{${h}_{2}$: Function estimating a steering angle}
\label{sec:h2} 

The function ${h}_{2}$ uses the following two assumptions:
\bulls{
\item In the three-dimensional (3D) vehicle coordinate system $(X,Y,Z)$, the $X$-axis points to the right, the $Y$-axis points forward from the vehicle, and the $Z$-axis points up from the ground. See the $X$- and $Y$-axes in Fig.~\ref{fig:turning radius}.
    
\item The vehicle movements along the $Z$-axis are ignorable.
}
Under the above two assumptions, ${h}_{2}$ estimates a steering angle using the vehicle pose at the $i$th and $(i-1)$th time point, ${\mb{P}}_{i,0}$ and ${\mb{P}}_{i-1,0}$ in \R{sys:vehicle's position}.
For simplicity, we transform the $i$th and $(i-1)$th vehicle poses in the world coordinate system to the vehicle coordinate system.

\subsubsection{Calculating a forward direction vector}
\label{sec:direct}

First, we calculate the forward direction vector of the vehicle at the $i$th time point.
The vehicle pose at the $i$th time point relative to that at the $(i-1)$th time point is given by
\be{
\label{sys:(i)-(i-1) pose}
\mb{P}_{i, i-1} 
= \mb{P}_{i,0} \mb{P}_{i-1,0}^{-1},
}
where we consider the vehicle at the $(i-1)$th time point is at the origin of a coordinate system, i.e., $\mb{P}_{i-1,0} = \mb{P}_{i-1,0} \mb{P}_{i-1,0}^{-1} = \mb{I}$.
Observing that
\bes{
\mb{P}_{i-1,0}^{-1} 
= 
\left[
\begin{array}{cc}
\mb{R}_{i-1,0}^{-1} & -\mb{R}_{i-1,0}^{-1} \mb{t}_{i-1,0}
\\
\mb{0}^\top & 1
\end{array}
\right],
}
we obtain the matrix in \R{sys:(i)-(i-1) pose} as follows:
\ea{
\label{sys:(i)-(i-1) pose:result}
\mb{P}_{i, i-1} 
= & 
\left[
\begin{array}{cc}
\mb{R}_{i,0} & \mb{t}_{i,0}
\\
\mb{0}^{\top} & 1
\end{array}
\right]
\cdot
\left[
\begin{array}{cc}
\mb{R}_{i-1,0}^{-1} & -\mb{R}_{i-1,0}^{-1} \mb{t}_{i-1,0}
\\
\mb{0}^\top & 1
\end{array}
\right]
\nn
\\
= & 
\left[
\begin{array}{cc}
\underbrace{ \mb{R}_{i,0} \mb{R}_{i-1,0}^{-1} }_{= \mb{R}_{i,i-1}}
& 
\underbrace{- \mb{R}_{i,i-1} \mb{t}_{i-1,0} + \mb{t}_{i,0}}_{= \mb{t}_{i,i-1}}
\\
\mb{0}^{\top} 
& 1
\end{array}
\right].
}
As we consider that $\mb{P}_{i-1,0} = \mb{I}$,
we now obtain the forward direction vector of the vehicle at the $i$th time point $\mb{d}_{i}$ by rotating the unit vector $[0, 1, 0]$:
\be{
\label{sys:forward vec}
\mb{d}_i = \mb{R}_{i,i-1}
\begin{bmatrix}
 0\\  1\\ 0
\end{bmatrix},
}
ignoring vertical movements as assumed above, 
where $\mb{R}_{i,i-1}$ is calculated in \R{sys:(i)-(i-1) pose:result}.
Fig.~\ref{fig:turning radius} shows the geometrical illustration of calculating $\mb{d}_i$ with the vehicle poses $\mb{P}_{i,0}$ and $\mb{P}_{i-1,0}$.
(Remark that it is inaccurate to calculate $\{\mb{d}_i\}$ from measures, e.g., direction and velocity, given by IMU(s), due to noise and bias in it/them \cite{IMU}.)

\subsubsection{Calculating a turning radius}
\label{sec:radius}

Second, using \R{sys:(i)-(i-1) pose:result} and \R{sys:forward vec}, we calculate the turning radius of the vehicle's front axle center (i.e., the radius of the vehicle's trajectory) at time step $i$.
Consider again that the vehicle pose at the $(i-1)$th time point is $\mb{I}$; see the $X$-$Y$ vehicle coordinate system in Fig.~\ref{fig:turning radius}.
In the $X$-$Y$ coordinate system, a line that is perpendicular to the directional vector $\mb{d}_i$ and crosses $\mb{t}_i$ is parameterized by
\bes{
Y = -\frac{d_{i,Y}}{d_{i,X}} X + \left( t_{(i,i-1),Y} + \frac{d_{i,Y}}{d_{i,X}} t_{(i,i-1),X} \right), 
}
where $d_{i,X}$ and $d_{i,Y}$ are the $X$- and $Y$-components of $\mb{d}_i$, respectively, and $t_{(i,i-1),X}$ and $t_{(i,i-1),Y}$ are the $X$- and $Y$-components of $\mb{t}_{i,i-1}$ (in \R{sys:(i)-(i-1) pose:result}), respectively.
We thus, obtain the turning radius of the vehicle at the $i$th time point (i.e., the $X$-intercept), $r_i$, as follows:
\be{
\label{sys:radius}
r_i = t_{(i,i-1),Y} \frac{d_{i,X}}{d_{i,Y}} + t_{(i,i-1),X}.
}
The sign of a turning radius indicates the turning direction: 
a positive and negative number indicates a right and left turn, respectively.
Fig.~\ref{fig:turning radius} illustrates the relations between the $i$th turning radius, $i$th forward vehicle vector, and the $i$th and $(i-1)$th vehicle poses in the $X-Y$ vehicle coordinate system. 

\begin{figure}[t!]
\centerline{\hbox{
\includegraphics[width=2.2in,trim={0.01cm 0.05cm 0.01cm 0.0cm},clip]{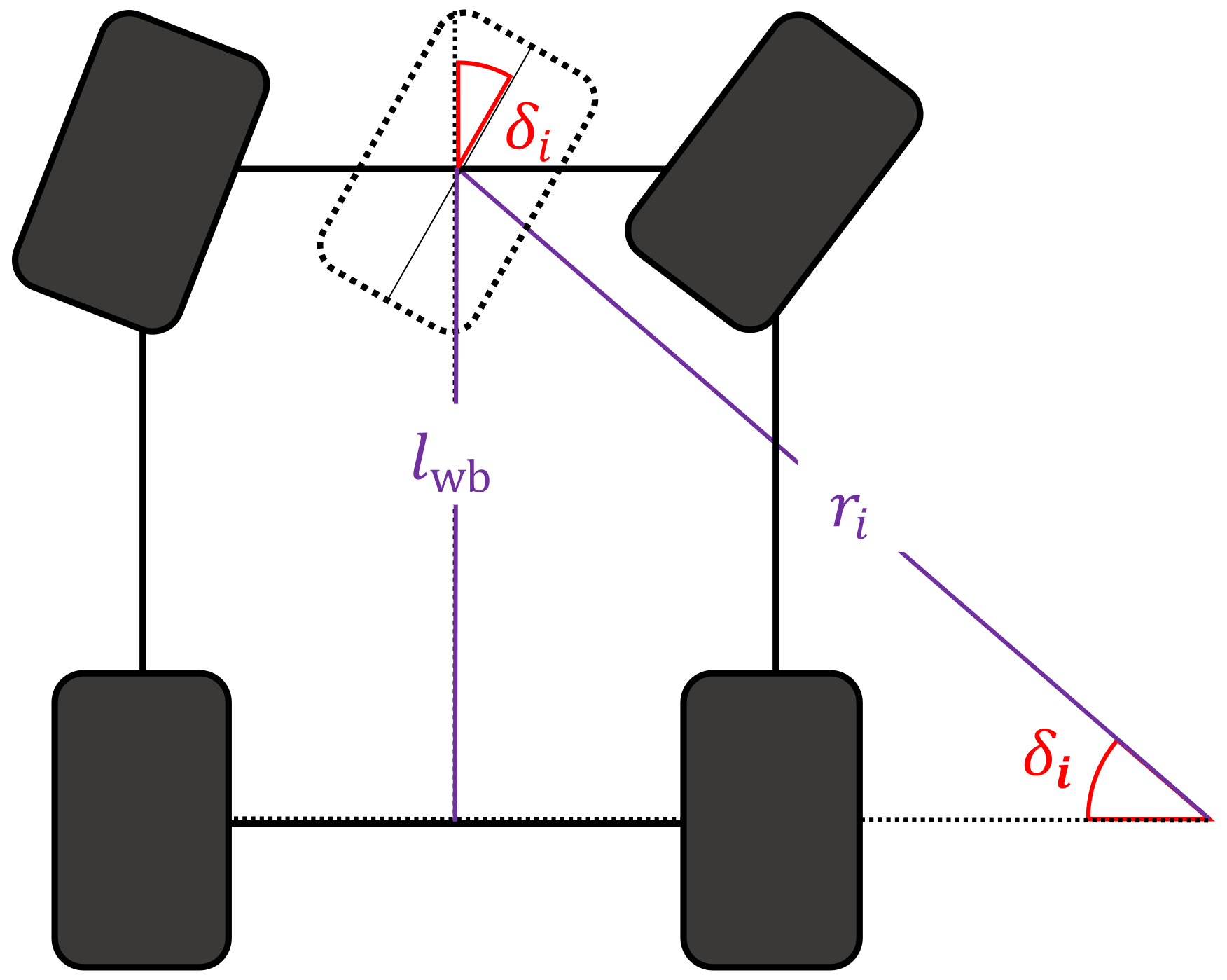}
}}
\vspace{-0.5pc}
\caption{The Ackermann steering geometry \cite{Vehicle_dynamics} for four-wheeled vehicles using front-wheel drive.
We set ${l}_{\text{wb}}$ as the actual wheelbase of the vehicle, and calculate $r_i$ as in \R{sys:radius}. 
In calculating $\delta_{i}$, we consider $l_{\text{wb}}$ and $r_i$ as the length of the opposite side and the length of the hypotenuse, respectively.}
\label{fig:parallel geometry}
\vspace*{-0.2in}
\end{figure}

\subsubsection{Calculating a steering angle}
\label{sec:angle}
Third, we estimate a steering angle for each time point using vehicle/steering geometry.
We use the calculated turning radius in \R{sys:radius} and two additional information about the vehicle, the wheelbase and steering ratio.
Throughout, we consider the Ackermann steering geometry, the most widely known steering geometry.\footnote{
We select the Ackermann steering geometry among three different types of steering geometries, Ackermann, anti-Ackermann, and parallel geometries.}
Fig.~\ref{fig:parallel geometry} illustrates the Ackermann steering geometry.
For simplicity of illustration, we consider front-wheel drive for four-wheeled vehicles. 

Let $l_{\text{wb}}$ be the wheelbase of the vehicle, the distance between the centers of the front and rear axles. Using the Ackermann steering geometry in Fig.~\ref{fig:parallel geometry}, we compute the degree of a front wheel turn $\delta_{i}$ at the $i$th time point as follows:
\be{
\label{sys:wheel angle}
\delta_{i}
= 
\arcsin \! \left( \frac{ l_{\text{wb}} }{ r_i } \right),
}
where ${r}_{i}$ is calculated as in \R{sys:radius}, and both $l_\text{wb}$ and $r_i$ are in the same unit.



There exists some difference between a steering wheel angle and a front wheel turn angle. 
This relation, so-called steering ratio, is a unique parameter of a vehicle:
\bes{
\text{steering ratio} = \frac{ y_i }{\delta_{i}}, 
}
where $y_i$ is the steering wheel angle at the $i$th time point, and $\delta_{i}$ is given as in \R{sys:wheel angle}.
Using this unique parameter of a vehicle and $\delta_{i}$ in \R{sys:wheel angle},
we finally obtain a pseudo steering angle $\hat{y}$ at the $i$th time point (see Fig.~\ref{fig:SSIL}) by
\be{
\label{sys:steering ratio}
\hat{y} = \text{steering ratio} \times \delta_{i}.
}

\subsection{$f$: E2E driving network from a camera, conditioned on high-level instructions}
\label{sec:networks}

\begin{figure}[t!]
\centerline{\hbox{
\includegraphics[width=2in]{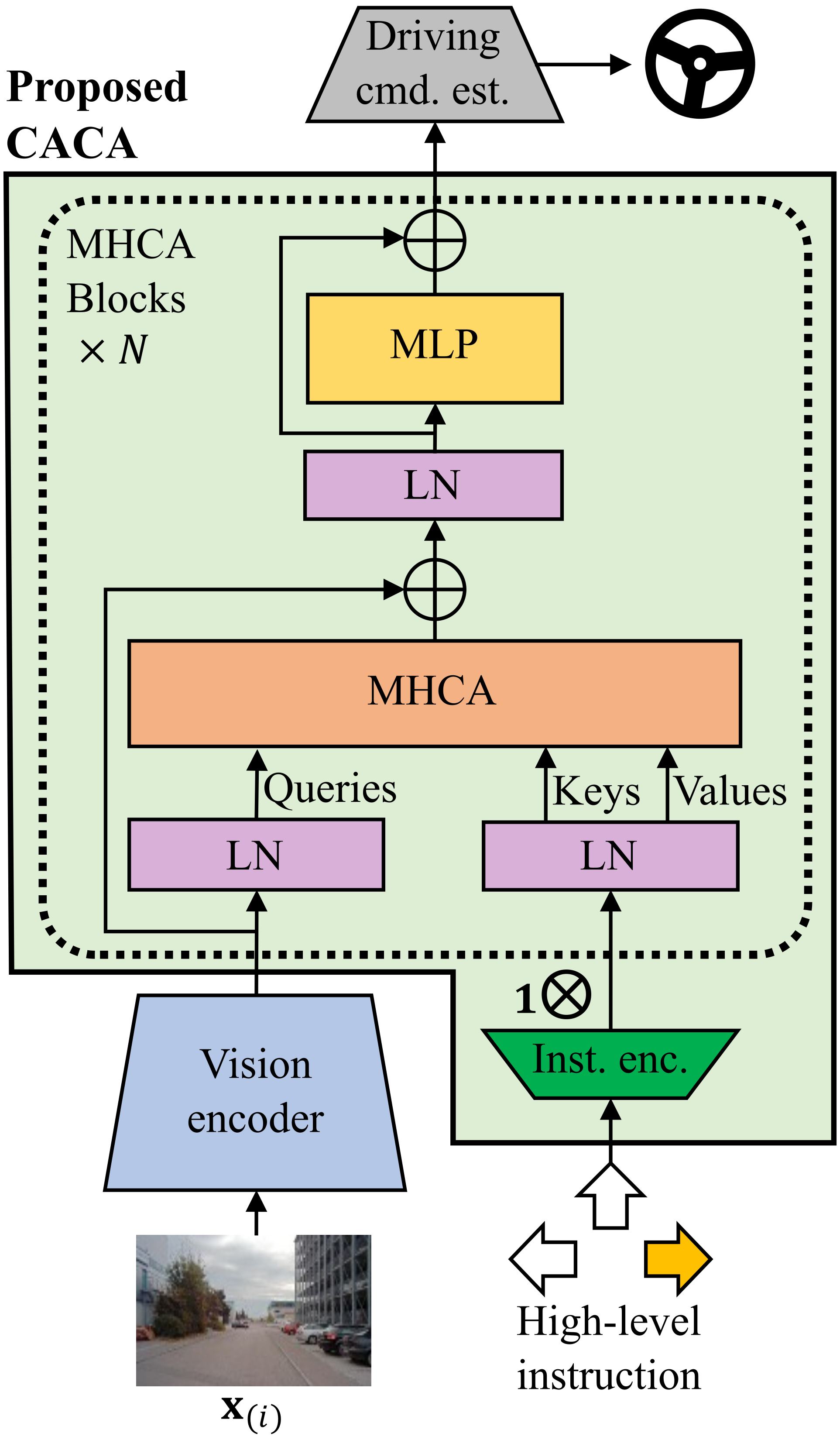}
}}
\vspace{-0.5pc}
\caption{
Our E2E network architecture using the proposed CACA.
LN denotes layer normalization and $\mb{1} \otimes (\cdot)$ denotes the Kronecker product between $\mb{1}$ and a column vector.
}
\label{fig:conditional_module}
\vspace*{-0.2in}
\end{figure}

Our aim is to learn driving policies leveraging vision information, conditioned on a high-level instruction command.
This section introduces a new vision-based E2E driving NN architecture $f$, 
by proposing a new conditioning approach, CACA, for a vision-based E2E driving network.

The proposed CACA mainly consists of $N$ multi-head cross-attention (MHCA) blocks.
In each MHCA block, 
we use extracted features by a vision-encoder backbone as the queries.
We encode a high-level instruction -- implemented by a one-hot vector -- through a single linear layer into an overcomplete representation,
and replicate (i.e., tile) this representation to construct the keys and values.
With the proposed design, we guide visual representations toward semantically relevant regions to improve E2E driving decisions.
Finally, we feed output features from the final MHCA block to a driving-command estimator implemented as a four-layer MLP with the number of neurons of $256$, $128$, $64$, and $1$.

In sum, using the pseudo labels in \R{sys:steering ratio}, we can train E2E driving NNs (in a self-supervised manner) that gives a steering angle for each time point from an image captured by a camera and a high-level instruction.
In obtaining the pseudo labels, we use point clouds from LiDAR sensor(s) and domain knowledge of the ego vehicle, including vehicle poses, steering geometry, wheelbase, and steering ratio.

\section{Experimental results and discussion}
\label{sec: experimental results and discussion}

This section compares the performances of vision-based E2E driving NNs trained by the ordinary SIL and the proposed SSIL framework, with three different benchmark datasets, two different vision-encoder backbones, and three different conditioning approaches.
In addition, 
Section \ref{sec:ablation_discussion} investigates different pseudo-label predictors in SSIL and Section \ref{sec:pretrain_discussion} investigates modified SSIL for feature learning with the perspective of the conventional self-supervised feature learning setup.

\subsection{Experimental setup: Datasets}
\label{sec:driving datasets} 
\subsubsection{A2D2}
\label{sec:a2d2}
We used A2D2 \cite{a2d2dataset} that for each frame, includes images collected from cameras, point cloud sets from LiDAR sensors, human driving commands, etc.
A2D2 consists of three driving scenes that were collected in different cities in Germany, Gaimersheim, Ingolstadt, and Munich.
The driving scenes from three cities include distinct road and lane characteristics in rural, suburban, and urban driving environments.
In each driving scene, we used 
\textit{1)} images collected from a camera at the center of the front header, 
\textit{2)}
point clouds from LiDAR sensors at the left, center, and right of the front header, \textit{3)} the human driven steering angle values, and \textit{4)} high-level instructions.
A2D2 provides high-level instruction commands 
at intersections, by following \cite{VAD}.
We changed the spatial resolution of input RGB images to $640 \!\times\! 160$; to better focus on road scene, we cut $60$ pixels from both the top and bottom.
For training, we used two driving scenes that include $12.6$K and $14.85$K frames collected from Gaimersheim and Ingolstadt, respectively.
For test, we used a driving scene that includes $15.75$K frames collected from Munich.

\subsubsection{nuScenes}
\label{sec:nuScenes}
The nuScenes dataset \cite{nuScenes} includes $1,\!000$ driving scenes collected in diverse weather conditions, times, and traffic environments.
For each frame, we used 
\textit{1)} an image collected from a front camera, 
\textit{2)} point clouds from a center LiDAR,
\textit{3)} steering angle values, and \textit{4)} high-level instructions by following \cite{VAD}.
We resized the RGB images resolution to $640 \!\times\! 160$ and cropped $60$ pixels from both the top and bottom, following the setup used in A2D2.
We divided the nuScenes dataset into $850$ and $150$ driving scenes that correspond to the total number of time points $1.19$M and $210$K, for training and test, respectively.

\subsubsection{CARLA}
\label{sec:carla}
We used the CARLA simulator with version $0.9.15$ to generate driving scenes in various simulation town environments.
In generating driving scenes, we followed the setup in \cite{Transfuser_1}.
\cite{Transfuser_1} provides the expert driving system of an ego-vehicle with sets of pre-defined routes and high-level instructions.
The generated driving dataset by an expert driving system includes \textit{1)} center camera images, 
\textit{2)} point clouds from a center LiDAR, 
\textit{3)} expert-driven steering angle values, and 
\textit{4)} high-level instructions for each frame.
We resized the input RGB images to $400 \!\times\! 300$ and cropped resized ones to $256 \!\times\! 256$ to reduce camera distortion.
For training, we used driving scenes collected from the five towns, Town$01$-Opt, Town$02$-Opt, Town$03$-Opt, Town$04$-Opt, and Town$06$-Opt.
For test, we used driving scenes collected from Town$05$-Opt.
They correspond to $365.9$K and $25.2$K time points, respectively.

\subsection{Experimental setup: Vehicle information}
\label{sec:vehicle_info}

The A2D2 dataset, the nuScenes dataset, and the simulated driving dataset by CARLA were collected by the Audi Q7 e-tron model, Renault Zoe, and Tesla Model 3, respectively.
The wheelbases $l_{\text{wb}}$ in \R{sys:wheel angle} are given as follows: for Audi Q7 e-tron, $l_{\text{wb}} = 2.994$m; for Renault Zoe, $l_{\text{wb}} = 2.924$m; for Tesla Model 3, $l_{\text{wb}} = 3.005$m.
The steering ratios of the Audi Q7 e-tron and the Renault Zoe vehicles are $15.8$ and $15.2$, respectively.
We did not use the steering ratio in  \R{sys:steering ratio} for the Tesla Model 3, as the CARLA simulator directly uses wheel turn angles instead of steering angles.

\subsection{Experimental setup: E2E driving NN architectures}
\label{sec:networks_setup}

Throughout performance comparison with different learning methods, driving datasets, and conditioning approaches,
we used two representative vision-based E2E driving network architectures, PilotNet \cite{PilotNet} (CNN) and Latent TransFuser \cite{Transfuser_2} vision transformer) as a vision-encoder backbone in Section~\ref{sec:networks}.
We compared the proposed CACA in Section~\ref{sec:networks} with two existing conditioning approaches (in E2E driving) using high-level instruction, 
the module selection approach \cite{E2E_Conditional_IL, E2E_Conditional_IL_2}, 
and the self-attention-based approach \cite{CIL++}.

\subsection{Experimental setup: Training}
\label{sec:traing}

We first finely tuned the training hyperparameters to obtain the best inference accuracy for SIL\footnote{The ordinary SIL setup trains E2E driving networks with pairs of RGB images and expert driving commands.} using the Latent Transfuser architecture in Section~\ref{sec:networks_setup}.
We set the initial learning rate, the batch size, and the number of epochs as $0.00001$, $512$, and $100$, respectively.
For CACA in Section \ref{sec:networks}, we used $N=4$.
We used four NVIDIA GeForce RTX 4090 GPUs throughout all the experiments.

We do not know the exact distribution of $\mb{x}$ in expected loss \R{sys:ssrl}, so as conventionally, we approximated it to the empirical SSIL loss function $\sum_{l=1}^L \| f( ( \mb{x}_{l} ) _{J^c} ) - g( (\mb{x}_l)_J ) \|_2^2$, where $(\mb{x}_l)_{J^c}$ and $(\mb{x}_l)_{J}$ are an image and point clouds at the $l$th frame, respectively, and $L$ is the number of training samples.
We pre-computed a pseudo steering angle for each frame by $\hat{y}_l = g( (\mb{x}_l)_J )$, for $l = 1,\ldots, L$, so that training times of SIL and SSIL models are identical, given the same NN architecture.

\subsection{Experimental setup: 
Pseudo-label predictors in SSIL}
\label{sec:ablation}

To investigate the contribution of \uline{different configurations of the designed pseudo-label predictor} (see Section~\ref{sec:ssil}), we generalized the previous time point index $i-1$ to $i-k$ in Section~\ref{sec:h2}, where $k$ denotes the temporal interval between two vehicle poses at the current time point and an arbitrary past time point.
We evaluated a pseudo-label predictor of SSIL with different $k$ values and corresponding E2E driving scores.
We set $k$ as $1$ (default), $2$, $4$, and $8$, and used Latent TransFuser using proposed CACA and the CARLA dataset.

To investigate the contribution of \uline{different types of pseudo-label predictors} in E2E driving performances,
we compared two pseudo-label predictors.
The first one is the proposed pseudo-label predictor in SSIL (see Section \ref{sec:ssil}); and the other one uses a PID controller.
For the PID controller-based pseudo-label predictor, 
we converted estimated vehicle poses to a steering angle using a PID controller, following \cite{Transfuser_1}.
We used Latent Transfuser using proposed CACA and the CARLA dataset;
and evaluated the estimation accuracy of 
the above two pseudo-label predictors and the corresponding E2E driving performances.

\subsection{Experimental setup: Extension of SSIL to self-supervised feature/representation learning}
\label{sec:pretrain_setup}

We modified the proposed self-supervised learning framework in Section~\ref{sec:ssil} with the perspective of the conventional self-supervised feature/representation learning setup that consists of (self-supervised) feature learning and (supervised) downstream task learning.
We followed its assumption that
an unlabeled dataset for self-supervised feature learning is large, while a labeled dataset for supervised downstream task learning is very small.
In feature learning, 
i.e., learning feature extractors, 
we used the nuScenes dataset (consisting of $1.19$M training samples) with pseudo steering angles obtained by SSIL.
In downstream task learning, i.e., learning MLPs, we used the Town01-Opt dataset of the CARLA dataset (consisting of $40$K training samples) with expert driving commands.
In SIL, we used the Town01-Opt dataset, following the supervised learning setup in \cite{Transfuser_2}.
Note that the two datasets use different vehicles.

\begin{table}[!t]
\captionsetup[table]{skip=6in}
\caption{E2E driving accuracy comparisons between SIL and SSIL with two different driving datasets, two different E2E driving NN architectures, and three different conditioning approaches (open-loop evaluation with MSE).}
\vspace{-0.5pc}
\centering
\setlength{\tabcolsep}{4pt} 
\renewcommand{\arraystretch}{1.1} 
\label{tab:recorded_dataset}
\begin{tabular}{c|cc|cc} 
\hline\hline
\multirow{2}{*}{\vspace{0.5mm}\begin{tabular}[c]{@{}c@{}} E2E driving \\ NN architectures \end{tabular}} 
& \multicolumn{2}{c|}{A2D2} 
& \multicolumn{2}{c}{nuScenes} 
\\ \cline{2-5}
& SIL & SSIL 
& SIL & SSIL  
\\ 
\hline
PilotNet (selection)
& 0.0264 & \textbf{0.0224} 
& \textbf{0.0094} & 0.0096
\\
Latent TransFuser (selection)
& \textbf{0.0084} & 0.0087
&  0.0058& \textbf{0.0056}
\\
PilotNet (self-attention)
& 0.0203 & \textbf{0.0199} 
& \textbf{0.0064} & 0.0065
\\
Latent TransFuser (self-attention)
& \textbf{0.0062} & 0.0063
& 0.0056 & \textbf{0.0053}
\\
\hline
PilotNet (our CACA)
& 0.0174 & \textbf{0.0157}
& 0.0054 & \textbf{0.0052}
\\
Latent TransFuser (our CACA)
& 0.0058 & \textbf{0.0057}
& \textbf{0.0046} & 0.0048
\\
\hline\hline
\end{tabular}

\end{table}

\begin{table}[t!]
\caption{E2E driving accuracy comparisons between SIL and SSIL with two different E2E driving NN architectures and three different conditioning approaches (closed-loop evaluation).}
\vspace{-0.5pc}
\centering
\setlength{\tabcolsep}{4pt}
\renewcommand{\arraystretch}{1.1}
\label{tab:simulator}
\begin{tabular}{c|cc|cc} 
\hline\hline
\multirow{3}{*}{\vspace{0.5mm}\begin{tabular}[c]{@{}c@{}} E2E driving \\ NN architectures \end{tabular}} 
& \multicolumn{4}{c}{CARLA} 
\\ \cline{2-5}
& SIL & SSIL 
& SIL & SSIL 
\\ \cline{2-5}
& \multicolumn{2}{c}{RC (\%)$^\uparrow$} 
& \multicolumn{2}{c}{DS (\%)$^\uparrow$} 
\\ 
\hline
PilotNet (selection)
& \textbf{18} & 17
& \textbf{14} & 13
\\
Latent TransFuser (selection)
& 50 & \textbf{51}
& \textbf{44} & 42
\\
PilotNet (self-attention)
& \textbf{19} & 18
& \textbf{15} & 14
\\
Latent TransFuser (self-attention)
& \textbf{53} & 52
& 43 & \textbf{44}
\\
\hline
PilotNet (our CACA)
& \textbf{24} & 22
& \textbf{18} & 17
\\
Latent TransFuser (our CACA)
& 62 & \textbf{64}
& 57 & \textbf{58}
\\
\hline\hline
\end{tabular}
\vspace{-0.15in}
\end{table}

\subsection{Experimental setup: Evaluation metrics}
\label{sec:metric}

To compare E2E driving performances between SIL and proposed SSIL, we used the following three metrics: \textit{1)} MSE, \textit{2)} route completion (RC), and \textit{3)} driving score (DS).
For open-loop evaluation that assesses E2E driving performance against pre-recorded driving behaviors, we used the conventional steering angle prediction measure MSE.
For closed-loop evaluation that assesses E2E driving performance using feedback from its previous driving decisions, 
we used 
RC defined by the percentage of the total driving distance completed, and
DS defined by the multiplication of the route completion and
the infraction score.
The DS metric can consider collision avoidance, on-road driving, etc.

\subsection{Comparisons between SIL and SSIL}
\label{sec:sil_ssil_discussion}

Regardless of driving datasets, E2E driving NN architectures, and conditioning approaches,
Tables~\ref{tab:recorded_dataset}--\ref{tab:simulator} demonstrate
that
\textit{without} using expert driving commands,
an E2E driving NN trained by proposed SSIL achieves \textit{very} comparable driving performances with that of ordinary SIL.
Considering that 
training and test sets of the A2D2 and CARLA datasets include different driving environments (see Section \ref{sec:driving datasets}),
the A2D2 and CARLA results in Tables~\ref{tab:recorded_dataset}--\ref{tab:simulator} show that 
given the same vehicle with the same sensor configurations,
trained E2E driving NNs via SSIL are well-transferable to new driving domains.

Regardless of vision-encoder backbones,
Tables~\ref{tab:recorded_dataset}--\ref{tab:simulator} show that the proposed CACA achieves significantly better driving performances compared to the existing conditioning approaches in E2E driving.
This implies that the proposed CACA can steer visual representations toward semantically relevant regions and thus improve driving decisions.

Throughout all experiments, we observed performance improvements by using the Latent TransFuser vision-encoder over PilotNet.
We think that using the attention mechanism with positional encoding in the Latent Transfuser can lead to a better understanding of contextual information than only using features from a CNN in PilotNet.

One could design an E2E driving NN $f$ to use the combination of camera(s) and partial LiDAR data, with a pseudo-predictor $g$ that uses partial LiDAR data.
More specifically, one could design $f$ to use 3D points from a set of LiDAR sensor(s) and $g$ to use those from the other LiDAR sensors, 
under the reasonable assumption that noises from different LiDAR sensors are statistically independent.

The SSIL times for the PilotNet and Latent Transfuser with the CACA are $3.12$ and $6.58$ minutes per epoch, respectively. 
Pseudo-label predictions in SSIL cost $31$ ms per a frame (with four NVIDIA GeForce RTX 4090 GPUs).

\subsection{Comparisons between different pseudo-label predictors in SSIL}
\label{sec:ablation_discussion}

The pseudo-predictor accuracy results and the corresponding E2E driving performances in Tables~\ref{tab:config}-\ref{tab:options} well-corresponds to Theorem~\ref{thm:ssrl} of SSRL in Section~\ref{sec:ssrl} implying that a better pseudo-predictor $g$ can lead to a more accurate regression network $f$.
 
In Table~\ref{tab:config}, we observed that a lower $k$ value, i.e., using a previous camera pose with more recent time, gives more accurate pseudo-label prediction, ultimately leading to better E2E driving performances.
To have better E2E driving performances via SSIL in dynamic driving environments, 
to use camera(s) and LiDAR sensor(s) with a higher sampling rate could be useful.
This is implied by the above result.

Table~\ref{tab:options} shows that 
compared to the existing PID-based pseudo-predictor, the proposed pseudo-label predictor gives more accurate pseudo-label prediction and better E2E driving performances.
In particular, using unique vehicle parameters, e.g., wheel based and steering ratio (see Section \ref{sec:h2}), in the proposed method can lead to accurate driving command prediction.

\begin{table}[t!]
\caption{Pseudo-label prediction accuracy and the corresponding E2E driving performances of SSIL with different temporal intervals ($k$) between two vehicle poses
(Pseudo-label prediction evaluation with MSE and closed-loop evaluation with RC and DS).}
\vspace{-0.5pc}
\centering
\setlength{\tabcolsep}{4pt}
\renewcommand{\arraystretch}{1.1}
\label{tab:config}
\begin{tabular}{c|c|ccc}
\hline\hline
\multirow{2}{*}{\begin{tabular}[c]{@{}c@{}} Temporal interval ($k$) \end{tabular}} 
& \multicolumn{3}{c}{CARLA} 
\\ \cline{2-4}
& \multicolumn{1}{c|}{MSE$^\downarrow$}
& \multicolumn{1}{c}{RC (\%)$^\uparrow$} 
& \multicolumn{1}{c}{DS (\%)$^\uparrow$} 
\\ 
\hline
$1$ (default)
& \textbf{0.00053}
& \textbf{64}
& \textbf{58}
\\
$2$
& 0.00075
& 37
& 28
\\
$4$
& 0.00096
& 14
& 5
\\
$8$
& 0.00190
& 6
& 2
\\
\hline\hline
\end{tabular}
\end{table}

\begin{table}[t!]

\caption{Pseudo-label prediction accuracy and the corresponding E2E driving performances of different types of pseudo-label predictors in SSIL
(Pseudo-label prediction evaluation with MSE and closed-loop evaluation with RC and DS).
}
\vspace{-0.5pc}
\centering
\setlength{\tabcolsep}{4pt}
\renewcommand{\arraystretch}{1.1}
\label{tab:options}
\begin{tabular}{c|c|ccc}
\hline\hline
\multirow{2}{*}{\begin{tabular}[c]{@{}c@{}} Pseudo-label prediction methods \end{tabular}} 
& \multicolumn{3}{c}{CARLA} 
\\ \cline{2-4}
& \multicolumn{1}{c|}{MSE$^\downarrow$}
& \multicolumn{1}{c}{RC (\%)$^\uparrow$} 
& \multicolumn{1}{c}{DS (\%)$^\uparrow$} 
\\ 
\hline
Proposed predictor
& \textbf{0.00053}
& \textbf{64}
& \textbf{58}
\\
PID controller-based predictor
& 0.00796
& 6
& 3
\\
\hline\hline
\end{tabular}
\end{table}

\begin{table}[t!]
\caption{E2E driving accuracy comparisons between different learning methods}
\vspace{-0.5pc}
\centering
\setlength{\tabcolsep}{4pt}
\renewcommand{\arraystretch}{1.1}
\label{tab:representation}
\begin{tabular}{c|c|c} 
\hline\hline
\multirow{2}{*}{\vspace{0.5mm}\begin{tabular}[c]{@{}c@{}} 
    Training methods
\end{tabular}} 
& \multicolumn{2}{c}{CARLA} 
\\ \cline{2-3}
& RC (\%)$^\uparrow$
& DS (\%)$^\uparrow$
\\
\hline
\begin{tabular}[c]{@{}c@{}}
    SIL (small labeled dataset)
\end{tabular}
& 13
& 8
\\
\begin{tabular}[c]{@{}c@{}}
    Modified SSIL for feature learning\\
    (large unlabeled dataset \& small labeled dataset)
\end{tabular}
& \textbf{24} 
& \textbf{13}
\\
\hline\hline
\end{tabular}
\vspace{-0.15in}
\end{table}

\subsection{Extension of SSIL to self-supervised feature/representation learning}
\label{sec:pretrain_discussion}

Table~\ref{tab:representation} shows that modified SSIL for self-supervised feature/representation learning outperforms ordinary SIL using expert commands.
The results imply that using a large unlabeled dataset with SSIL can learn more informative features than using only a small labeled dataset with SIL.


\section{Limitations}
\label{sec:limitations}

This section discusses three limitations of the proposed SSIL framework.
First, SSIL depends on the performance of SLAM for pseudo-label prediction.
The proposed method in Section~\ref{sec:ssil} uses a SLAM method using LiDAR data, A-LOAM \cite{LOAM}.
Its performance may degrade in degenerative environments, 
e.g., when LiDAR faces a single texture-less wall, and severe weather conditions, e.g., heavy rain, fog, and snow.
In such feature-scarce environments, it is difficult for LiDAR-based SLAM to perform accurate the LiDAR sensor odometry and scan matching.
This is supported by our observations that compared to the averaged pseudo-label prediction accuracy in Table~\ref{tab:config} ($k\!=\!1$), that in degenerative environments is lower, specifically, $\text{MSE} = 0.0044$.

Second, a trained E2E driving NN by proposed SSIL using domain knowledge of a vehicle (i.e., the wheelbase in \R{sys:wheel angle} and steering ratio in \R{sys:steering ratio})
may be only suboptimal for vehicles with different vehicle parameters.
Because the wheelbase \R{sys:wheel angle} and steering ratio \R{sys:steering ratio} are intrinsic vehicle parameters, changing them to different ones needs to re-predict pseudo labels for the SSIL of an E2E driving NN.

Third, 
similar to existing E2E driving methods using high-level instructions \cite{E2E_Conditional_IL, E2E_Conditional_IL_2, CIL++},
if inaccurate high-level instructions are given, we expect performance degradations in the proposed framework.

\section{Conclusion}
\label{sec:conclusion}

To train accurate and reliable E2E driving networks, it is crucial to construct vehicle control datasets.
However, it is extremely challenging to access such vehicle data, e.g., steering angle control, \textit{without} the assistance of vehicle manufacturers.
To address this challenge in vision-based E2E driving, we propose the first SSIL framework that can predict pseudo steering angles using LiDAR sensor(s) and vehicle information.
The proposed pseudo-label predictor outperformed existing one using PID controller.
Moreover, the proposed CACA achieved superior driving performances over the existing conditioning approaches using high-level instructions.
Our experimental results with three different benchmark datasets demonstrate that the proposed SSIL framework can achieve \textit{very} comparable performance with that of the ordinary supervision counterpart.

Our first future work is to extend the proposed SSIL framework to construct pseudo labels for other driving commands, such as velocity and brake throttle, using the proposed vehicle forward direction estimation in Section~\ref{sec:direct}.
Our second future work is to overcome the limitation of vision-based E2E driving NNs by developing multi-modal E2E driving NNs that use both camera and LiDAR data \cite{Transfuser_1} to compensate for weaknesses of each sensor, possibly at the cost of computational complexity.
We will design effective pseudo-label predictors and E2E driving networks for SSIL to estimate more diverse driving commands and be robustly adaptable to various driving scenarios, e.g, unexpected object appearing, an unprotected right turn at an intersection, etc., and some possible errors in high-level instructions. 

\bibliographystyle{IEEEtran}
\bibliography{referenceBibs_AD_JB}

\end{document}